\documentclass[runningheads]{llncs}

\usepackage{eccv2024style/eccv}

\usepackage{eccv2024style/eccvabbrv}

\usepackage{graphicx}
\usepackage{booktabs}

\usepackage[accsupp]{axessibility}  % Improves PDF readability for those with disabilities.

\usepackage{hyperref}

\usepackage{orcidlink}

\newcommand{\lastvisiteddatum}{6.8.2024}

\begin{document}

\title{Art Forgery Detection using Kolmogorov Arnold and Convolutional Neural Networks } 

%\title{Art Forgery Detection of Wolfgang Beltracchi } 

\titlerunning{Art Forgery Detection of Wolfgang Beltracchi}

\author{Sandro Boccuzzo\inst{1}\orcidlink{0009-0000-7538-7068} \and
Deborah Desirée Meyer\inst{1}\orcidlink{0009-0007-6557-4661} \and
Ludovica Schaerf\inst{2}\orcidlink{0000-0001-9460-702X}
}

\authorrunning{S.~Boccuzzo et al.}

\institute{University of Zurich (UZH), Rämistrasse 71, 8006 Zürich, Switzerland
\email{\{sandro.boccuzzo,deborahdesiree.meyer\}@uzh.ch}\\
\and
University of Zurich (UZH), Digital Visual Studies (DVS), Culmannstrasse 1, 8006 Zürich, Switzerland\\
\email{ludovica.schaerf@uzh.ch}}

\maketitle

\begin{abstract}
  Art authentication has historically established itself as a task requiring profound connoisseurship of one particular artist. Nevertheless, famous art forgers such as Wolfgang Beltracchi were able to deceive dozens of art experts. In recent years Artificial Intelligence algorithms have been successfully applied to various image processing tasks. In this work, we leverage the growing improvements in AI to present an art authentication framework for the identification of the forger Wolfgang Beltracchi. Differently from existing literature on AI-aided art authentication, we focus on a specialized model of a forger, rather than an artist, flipping the approach of traditional AI methods. We use a carefully compiled dataset of known artists forged by Beltracchi and a set of known works by the forger to train a multiclass image classification model based on EfficientNet. We compare the results with Kolmogorov Arnold Networks (KAN) which, to the best of our knowledge, have never been tested in the art domain. The results show a general agreement between the different models' predictions on artworks flagged as forgeries, which are then closely studied using visual analysis. 
\end{abstract}

\section{Introduction}
\label{sec:introduction}
\noindent As objects of cultural significance, the value of artworks is inherently tied to their authenticity. This is reflected in the sale of works on the art market. It is integral to the daily practice of auction houses, for instance, to verify the authenticity of a work, in order to accept, value, and sell it on the market. This is based on visual and contextual elements. The former include the material qualities, technical skills, and restoration status of a work. The latter encompasses its alignment with an artist's oeuvre, historical context, provenance history, and the value of comparable works. As such, an artwork's authenticity and concomitant value are determined. Momentary reflections of artistic value are captured in auction records, interacting with current socio-cultural and economic factors. Here, the value ascribed to authentic works becomes seemingly unbounded by financial constraints, reaching ever-increasing sales records. With authenticity at its core, an artwork thus bears cultural value and generates capital. This incentivizes the forgery and exploitation of objects of cultural heritage.

\indent Yet the authentication of artworks is not always a straightforward endeavor. This is exemplified by the notorious art forger, Wolfgang Beltracchi, and the continuous circulation of his forgeries on the art market. Beltracchi has forged hundreds of paintings of 20th-century artists such as Max Pechstein, Heinrich Campendonk, Kees van Dongen, and Max Ernst. While released after three years of prison in 2015, only 14 of around 300 forgeries have been determined imitations of original paintings and their styles\footnote{\url{https://news.artnet.com/art-world/master-forger-wolfgang-beltracchi-released-from-prison-225025} (\lastvisiteddatum).}.  As such, accurate and efficient methods for detecting Beltracchi's forgeries must be employed to minimize the risk introduced by the circulation of his works on the art market.

\indent While technologies have been used to support the discovery of forged paintings, using methods such as infrared light, X-ray, carbon dating, or pigment analyses, the process of art authentication has greatly developed with the application of Artificial Intelligence algorithms to various image processing tasks. These approaches aim to verify the authenticity of genuine artworks created by established artists. Flipping the approach of traditional methods, this paper seeks to tackle art authentication from the perspective of the forger, detecting forgeries made by a known art forger. As an artist's hand can be distinguished by looking at the visual elements of a painting, potential subtleties in a forger's approach to painting may be detected by the use of AI. This becomes particularly interesting in the case of Wolfgang Beltracchi, where specific forgeries and forged artists are known, and a large section of the oeuvre is still undetected.

\indent In this paper, we focus on the detection of paintings forged by Wolfgang Beltracchi using computer vision models. We create an ad hoc dataset of artists known to be forged by Beltracchi alongside a set of recognized works by the forger. We train a multiclass image classification model using EfficientNets and a Kolmogorov Arnold Network (KAN). 
Differently from the existing literature, our aim is not to directly detect Beltracchi's forgeries, but rather to signal possible undiscovered forgeries in the dataset and investigate, through art historical analyses, the validity of the attribution. As the majority of Beltracchi's forgeries of the artists in consideration are undiscovered, we assume a mislabeling of some images in our authentic set. We train several splits of the dataset, to be able to test the whole dataset and mitigate the influence of mislabeled data in the training. We then analyze the artworks whose patches are most disputed according to our methods. We identify the stylistic features that the algorithm links with concerns of authenticity and thus raises as signs of forgery. 
\newpage
In sum, the contributions of the paper are the following:
\begin{enumerate}
    \item We propose a variety of pre-processing steps to reduce digital image noise and test their effect,
    \item We investigate the influence of model size on art authentication performance,
    \item We compare results on CNNs with the newly introduced KAN model. To the best of our knowledge, KAN has never been applied to the arts, 
    \item We propose a framework to handle mislabeling of the training data through computer vision and art historical analyses. 

\end{enumerate}

The remainder of this paper is organized as follows. 
In Section \ref{sec:relatedwork}, we address related works on art authentication using artificial intelligence, in order to contextualize our work in the broader field. 
In Section \ref{sec:methodology} 
we discuss our methodology, introducing the preprocessing and models.   
In Section \ref{sec:data} the dataset and parameters used to train the model are discussed. In Section \ref{sec:evaluation} we evaluate our trained models. In Section \ref{sec:results} we present the results of the potential forgeries detected in our tests, concerning their statistical analysis and art historical interpretation. In Section \ref{sec:criticalDiscussion} we address the limitations of our approach. Finally, our findings and considerations for future work are summarized in Section \ref{sec:conclusions}.

\section{Related Work}
\label{sec:relatedwork}

The complex task of art authentication can greatly benefit from interdisciplinary support, moving beyond the art historical domain of visual analysis and the scientific application of technical devices. Numerous approaches to using Artificial Intelligence for the authentication of artworks have been made to this effect. These include the combination of AI with the analysis of physical attributes and extend to the use of exclusively digital images.

Exemplary of such a combinatory approach is the use of optical coherence tomography to analyze the morphology of cracks in original versus forged paintings \cite{kim2022investigation}. The majority of approaches, however, focus on the use of digital images. These vary from the use of attribute markers in machine learning algorithms trained on images of various artists' contemporary paintings \cite{dobbs2023contemporary}, vision transformers on a dataset of Vincent van Gogh paintings with the use of two different contrast sets \cite{schaerf2023art}, synthetic images to enrich the set of forgeries \cite{ostmeyer2024synthetic}, residual neural networks on the Rijksmuseum dataset \cite{dobbs2022art}, and the use of convolutional neural networks, trained on Raphael and comparable works to detect forgery and determine attribution probabilities \cite{frank2022Raphael}, or a collection of artworks for automatic artist attribution according to artist-specific visual features \cite{VanNoord2015}. Several studies also focus on the use of patches, using multiple patches to verify the creation of various works by the same artist \cite{jung2023patches}, or using Transformers applied to image patches for image classification \cite{dosovitskiy2021image}. Moreover, a case study by the company Art Recognition uses a Max Pechstein algorithm used to analyze brushstrokes, detecting Wolfgang Beltracchi's forgery of his \textit{Seine Bridge with Freight Barges}.\footnote{\url{https://art-recognition.com/case-studies/max-pechstein-seine-bridge-with-freight-barges} (\lastvisiteddatum).}

While applicable to forgery detection, these approaches are collectively trained on, and analyze, the authenticity of artworks created by specific established artists. By verifying the authenticity of works based on the artists' hands, forgery detection is positioned as a side-effect. Alternatively, we focus on the direct detection of forgery by learning the stylistic approach of the known art forger, Wolfgang Beltracchi.

% \cite{van2015toward}
% \cite{bell2021reflections}
% \cite{zhu2019machine}

% \footnote{https://upload.wikimedia.org/wikipedia/commons/1/10/REPORT_example_Cezanne_Boy_in_a_red_vest_ENGLISH._ENCRYPTED.pdf (\lastvisiteddatum).}

%Since the release of AlexNet
%
%Convolutional neural networks gained considerable popularity with the 2012 release of AlexNet, 
%
%which largely outperformed all previous models at the ILSVRC Ima- geNet Challenge 2012 [12, 25, 26].
%
%EfficientNet are two of the most successful CNNs. ResNet, as described by He et al. [23], is a (po- tentially) extremely deep CNN. In contrast to a standard CNN, where each stage learns a function F(x) based on input x, ResNet stages learn the residual function   x by using skip connections. The use of skip con- nections allows ResNets to excel due to their increased depth. EfficientNet represents a class of CNN models introduced by Tan and Le [24]. These models are opti- mized by scaling the width, depth, and input resolution of CNNs with a fixed ratio. EfficientNets have demonstrated superior performance compared to ResNets on image classification tasks. In our experiments, we use the variants ResNet-101 and EfficientNetB5.

\section{Methodology}
\label{sec:methodology}

AI-aided art authentication is often tackled as a binary image classification task, with the class 0 belonging to the artist one wishes to identify, and the class 1 indicating what is opposed to the artist. In the case of an artist-based authentication, the second class would include forgers, imitators, and similar artists. In our case, we consider one forger and several authentic artists. Due to the different nature of each artist, and because we wish to conduct a direct comparison between the artists and forger, we tackle the authentication as a multiclass classification task, where each class is a genuine artist forged by Beltracchi and a class for the forger.

\subsection{Pre-processing}
The paintings in all classes are split into center-cropped patches of $256 x 256$ pixels. Splitting paintings into smaller patches provides a simple but effective data augmentation, allowing the model to focus on more details when learning, this approach has been used in other works such as \cite{schaerf2023art}. 

We notice that the resulting patches contain some superficial artifacts, due to the digital encoding or acquisition setting. As this problem might hinder the training, we test the application of a blurring filter to the patches, to create an image that is less hazy than the original one. We use a Gaussian blur. This technique has been used in various other fields such as darkroom photography, or digital image processing and research such as \cite{padole2015improved}. 

Furthermore, some of the patches contain very minimal information (i.e. patches containing background), which could yield arbitrary predictions. We use the entropy measure to quantify the amount of information contained in each patch. We calculate the entropy of each individual color channel (RGB) separately for each patch and take the average of the three entropies. We test with different thresholds below with which we filter out uninformative patches.

\subsection{Models}
In this work, we compare state-of-the-art CNN models for art classification (Efficientnets) to the newly introduced paradigm of Kolmogorov Arnold Models (KAN). 

\subsubsection{Efficientnets}
EfficientNet is a family of convolutional neural networks that scale efficiently with depth, width, and resolution, introduced by Tan and Le \cite{tan2019efficientnet}. EfficientNet utilizes a compound scaling method that uniformly scales all dimensions of depth, width, and resolution using a simple yet effective formula. This model architecture has proven to be highly effective in achieving competitive results across various image classification benchmarks, including achieving state-of-the-art performance in art authentication \cite{schaerf2023art}.

In the art classification and attribution domain, as suggested by Elgammal \etal, it is unclear whether a deeper network will necessarily perform best \cite{elgammal2018shape}. In our work, we test this intuition by training the same dataset with different versions of different sizes of EfficientNet \cite{tan2019efficientnet} and comparing the results.

To adapt Efficientnet pre-trained on Imagenet to our task, we substitute the last layer with $12$ nodes with softmax activation and finetune the entire model using cross-entropy loss and Adam optimizer. 

\subsubsection{Kolmogorov Arnold Network (KAN)}

Kolmogorov-Arnold Networks (KAN) are models based on the Kolmogorov-Arnold representation theorem, which states that any multivariable continuous function can be decomposed into a finite sum of continuous non-linear functions of one variable \cite{liu2024kan}. In the context of neural networks, KAN models leverage this theorem to break down complex, high-dimensional data using B-splines as learnable activation functions and summing the nodes. 

To adapt KAN to our model, we instantiate a three-layer model with 12 nodes in the last layer\footnote{We use the implementation available here: \url{https://github.com/chinmayembedded/KAN\_for\_image\_classification} (\lastvisiteddatum).}. We train using cross-entropy loss and SGD optimizer. We test the cubic polynomial order of the splines.

\section{Data}
\label{sec:data}

Our dataset consists of 11 artists known to have been forged by Beltracchi and the forger himself. The list of artists can be found in Table \ref{table:artistInDataset}, alongside the artwork counts and patch numbers. To gather the images of the known artists, we use Wikiart, a publicly available dataset containing around 80K images. At times, the quality of the Wikiart images is poor and in a compressed file format. The quality of image content in art authentication is essential, therefore, we reduced the dataset to images that are at least 768 pixels wide. We also focus on paintings, thus excluding etching, sculptures, etc. 

We collect $53$ paintings by Beltracchi using various sources. We include artworks from the forger's personal website containing legitimate imitations and the forgeries listed in the official publication from the trial. The paintings contain both the identified forgeries, some imitations of other artists, and original artworks in his repertoire.

The filtered dataset has a total of $1334$ paintings. $134$ paintings ($10\%$) are allocated to the test set. The remaining $90\%$ paintings were split into a training set of $960$ ($80\%$) and a validation set of $240$ ($20\%$). The division into training, validation, and test sets maintains the distribution of artists largely constant (Table \ref{table:artistInDataset}). We repeat the split using 10 random seeds, obtaining 10 dataset versions. In what follows, we use the random seed number to name the datasets. We ensure that each artwork by the 11 artists appears at least once in the test set. Together with providing statistical relevance to our results, the different splits allow testing on all artworks currently deemed authentic, thus being able to flag possible unrecognized forgeries.

\begin{table*}
% \begin{table}
  \caption{Artists and the corresponding number of paintings and patches in the dataset.}
  \label{table:artistInDataset}
  \centering
  \begin{tabular}{lccc}
    \toprule
    Artist & Paintings |& In training  |& Patches \\
    \midrule
    Vincent van Gogh & 420 & 305 & 27021  \\
    William Turner & 163 & 117 & 5028 \\
    Gustav Klimt & 109 & 81 & 6331 \\
    Max Ernst & 99 & 67 & 2282 \\
    Fernand Leger & 106 & 76 & 4537 \\
    Andre Derain & 88 & 61 & 2491	\\
    August Macke & 81 & 59 & 4547	\\
    Maurice de Vlaminck & 67 & 50 & 270 \\
    Jean Metzinger & 65 & 46 & 2288 \\
    Wolfgang Beltracchi & 53 & 36 & 1368 	 \\
    Max Pechstein & 42 & 31 & 1922 \\
    Heinrich Campendonk & 41 & 27 & 1587 \\
    \midrule
    Total & 1334 & 960 & 59672 \\
    \bottomrule
  \end{tabular}
\end{table*}

%\begin{figure*}
 % \centering
  %\includegraphics[width=\linewidth]{images/verteilung}
  %\caption{Represented artists and paintings in the dataset.}
   %\label{figure:representedArtists}
%\end{figure*}

%\noindent 

\section{Evaluation}
\label{sec:evaluation}

In this section, we evaluate the two pre-processing techniques, determine the optimal model size, and compare CNN's Efficientnet to KAN. 

\subsubsection{Entropy comparison}

In this subsection, we evaluate the entropy-based patch filtering method. 
For this, we train EfficientNet B0 on the 10 training set splits with entropy-based filtering of 0, 2.5, and 3. We compare the average accuracy of the validation sets. The comparison shows an improvement by the model by filtering out patches with lower information (Table \ref{table:entropyComparison}). 
 The difference between the results is not significant. Nonetheless, we find that for our dataset the highest accuracy across different training set selections is obtained using an entropy threshold of 2.5.

\begin{table*}
  \caption{Entropy comparison on the datasets $T38-T43$ with EfficientNet B0}
  \label{table:entropyComparison}
  \centering
  \begin{tabular}{lcccr}
    \toprule
    Threshold | & Patches train. | & Patches valid. | & Min. valid. loss | & Avg. val. acc. (std)\\
    \midrule
    No filtering & 44685 & 9076 & 0.6349 & 77.3\% (2.45)  \\
    Entropy > 2.5 & 43127 & 8844 & 0.6595  & \textbf{77.4\%} (2.90) \\
    Entropy > 3 & 42641 & 8783 & 0.6288 & 77.2\% (3.12) \\
    \bottomrule
  \end{tabular}
\end{table*}

We therefore use this filtering in the following experiments.

\subsubsection{Blurring comparison}

In this subsection, we evaluate the blurring filter. Similarly to the previous test, we train EfficientNet B0 on 10 splits and maintain the entropy filtering of 2.5. Once on the original images and once on the images that have been slightly blurred. We compare their average accuracy on the validation set and observe a slight, non-significant, improvement using blurring (Table \ref{table:datasetComparison}). 

\begin{table*}
  \caption{Dataset comparison on the datasets $T38$-$T47$ with EfficientNet B0 and entropy > 2.5.}
  \label{table:datasetComparison}
  \centering
  \begin{tabular}{lcccr}
    \toprule
    Dataset & Patches train. | & Patches valid. | & Min. valid. loss | & Avg. val. acc. (std)\\
    \midrule
    Original & 43127 & 8844 & 0.6176 & 76.8\%  (3.32)\\
    Blur filtering & 43127 & 8844 & 0.6295 & \textbf{77.4\%} (2.90) \\
    \bottomrule
  \end{tabular}
\end{table*}

\subsubsection{Model size comparison}

\begin{figure*}[ht]
  \centering
    \includegraphics[width=8cm]{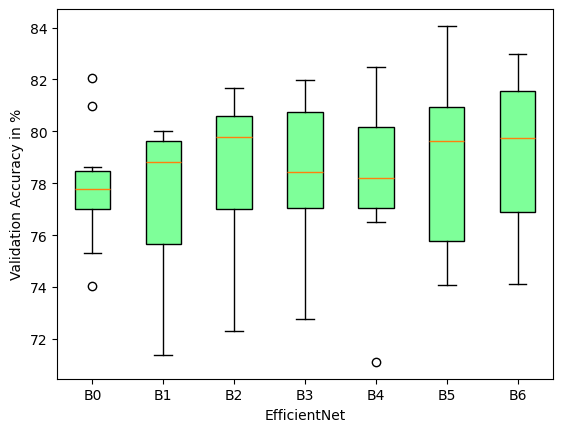}
  \caption{Validation accuracy in \% over EfficientNet $B0$ - $B6$ using dataset $T38$ - $T47$ and entropy > 2.5.}
   \label{figure:efficientnet}
\end{figure*}

In this subsection, we evaluate the selection of the EfficientNet version used as the base model for our work. For this, we train our model versions from 1 to 6 on the 10 training splits with an entropy threshold of 2.5 and a blur filter. We compare their average accuracy on the validation set. The comparison shows that, as suggested by Elgammal \etal \cite{elgammal2018shape}, a deeper network in our domain does not result in necessarily better performance. For our domain, we encounter that the complexity of EfficientNet B2 achieves the highest accuracy (Figure \ref{figure:efficientnet}). However, a good tradeoff between accuracy and computation costs can be found already in Efficientnet B0.

\subsubsection{Comparison with KAN}
Lastly, we train a simple KAN architecture with three or four layers and varying widths, using a grid size of 5 and a spline order of 3. We show the average validation accuracy on the $T38$ dataset in Table \ref{table:KAN}. We observe that a medium-sized model of $120x32x12$ achieves the best performance for KAN. The accuracies are lower than the established EfficientNet, however, we could not sufficiently explore the parameter space to determine the superiority. 

\begin{table*}
  \caption{KAN accuracies on the datasets $T38$. }
  \label{table:KAN}
  \centering
  \begin{tabular}{llccr}
    \toprule
    Grid size|& Patches train. | & Patches valid. | & Min. valid. loss | & Val. acc.\\
    \midrule
    24|16|12 & 43127 & 8844 & 1.6138 & 51.0\% \\
    64|32|12 & 43127 & 8844 & 1.8566 & 46.7\% \\
    120|84|12 & 43127 & 8844 & 1.6726 & \textbf{52.8\%} \\
    128|64|32|12 & 43127 & 8844 & 1.8421 & 46.3\% \\
    256|128|64|12 & 43127 & 8844 & 1.8829 & 45.3\% \\
    768|256|12 & 43127 & 8844 & 1.9244 & 46.5\% \\
    \bottomrule
  \end{tabular}
\end{table*}

\newpage
\subsubsection{Flagging questionable patches}
In this subsection, we discuss the issue of selecting an authentic training set from a pool including potential forgeries. As introduced previously, more than 250 forgeries by Beltracchi have likely not yet been identified. This introduces a high risk of training on fake artworks as authentic. The issue of introducing fake artworks into the original dataset (label noise) has also been discussed by Schaerf \etal \cite{schaerf2023art, ostmeyer2024synthetic}. 
To address this issue, we train models with different training selections and compare the predictions of all patches when appearing in the test set (Table \ref{table:patchesDetection}). We observe that a constant number of paintings is detected as possible forgery, this is roughly $1\%$ of the data. This result is encouraging as the number seems low enough not to confuse the model heavily. In the following sections, we consider predictions on artworks when appearing in the test set of each configuration for analyses.

\begin{table*}
  \caption{Detected patches/paintings with training sets $T38$ - $T47$ using EfficientNet B0 and entropy > 2.5.}
  \label{table:patchesDetection}
  \centering
  \begin{tabular}{lcccccccccc}
    \toprule
    Originals predicted as forgeries |& $T38$ & $T39$ & $T40$ & $T41$ & $T42$ & $T43$ & $T44$ & $T45$ & $T46$ & $T47$ \\
    \midrule
     Patches &
      151 & 180 & 184 & 54 & 51 & 120 & 82 & 116 & 94 & 101 \\
    Paintings &
      18 & 21 & 20 & 13 & 16 & 17 & 14 & 17 & 19 & 17 \\
    \bottomrule
  \end{tabular}
\end{table*}

\section{Results}
\label{sec:results}
\subsection{Statistical Analysis}
%\noindent 
In this subsection, we take a closer look at the results obtained by the models in detecting potential forgeries.

To evaluate the patches attributed to Beltracchi, we considered the patches with the 20 highest attribution values calculated by the model as Wolfgang Beltracchi, when appearing in the test set.

We observe that some patches remain coherently predicted as Beltracchi between EfficientNet B0 and KAN. This is a good indication of the validity of the predictions. Some paintings exhibit only one or a few patches detected by only one particular model as forgeries. Even though it is possible that only one area in one particular patch shows Beltracchi's hand, in this work, we focus on the paintings that have more than 1 patch misattributed.   

We show 21 paintings that appear among the most strongly attributed Beltracchi for either Efficientnet B0 or KAN, shown in Table \ref{table:patchesDetectionT38T47artists}.

\begin{table*}
  \caption{Paintings with more than 2 patches in the \textit{Top20} (\textit{classified as Beltracchi}) of every dataset $T38$ - $T47$ with EfficientNet B0 and on dataset $T38$ for KAN (Width 64/32/12, Grid = 5, k = 3) and entropy > 2.5. As KAN was trained only on $T38$ we indicate by $-$ when the image does not appear in the test set.}
\label{table:patchesDetectionT38T47artists}
  \centering
  \begin{tabular}{llcc}
    \toprule
Artist & Painting & Avg. Patches & Patches\\
  &   &   Efficientnet B0 & KAN\\
\midrule
Andre Derain & Red Sails & 6 & -  \\
 & London Bridge & 4 &  - \\
 & Fishing boats in Collioure & 0 &  2 \\
 & Two Barges & 3 & - \\
 & Banks of Seine & 4 & 7 \\
 & Still life with pears & 5 & 1 \\
 & and indian bowl & &  \\
August Macke & St. George & 9 & - \\
Gustav Klimt & Ode To Klimt & 6 & 0 \\
Heinrich Campendonk & Girl with chicken & 16 & - \\
 & Cyclist E. \& little yellow cow & 1 & 1 \\
Jean Metzinger & Corkscrew glass and matches & 3.5 & 0 \\
 & The Card Party & 9 & - \\
 & Coffee grinder with stemmed & 0 & 1 \\
 & glass and tea box &  &  \\
 & The path through the fields & 10 & - \\
 & The Harlequins & 10 & - \\
 & Woman with a mandolin & 3 & - \\
Max Pechstein & Fishing boats in Nidden & 7.5 & - \\
 & Evening clouds & 0 & 1 \\
 & Cutter in the millrace & 0 & 3 \\
Vincent van Gogh & Swift & 3.5 & - \\
William Turner & Venice, The Dogana  & 6 & -\\
& San Giorgio Maggiore &  &  \\
    \bottomrule
  \end{tabular}
\end{table*}

%\begin{table*}
 % \caption{Detected patches attributed to original artists predicted as on average Beltrachi > 0.5.}
 % \label{table:patchesDetectionT38artistsDifferenceGreater3}
  %\centering
  %\begin{tabular}{llcccccccc}
  %  \toprule
   % Artist & Painting & B0 & B1 & B2 & B3 & B4 & B5 & B6 & KAN \\
   % \midrule
%André Derain & Banks of Seine & 10 & 6 & 1 & 19 & 2 & 18 & 8 & 7\\
% & London Bridge & 15 & 5 & 6 & 3 & 14 & 8 & 0 & 0\\
 %& Still Life with Pears & 5 & 5 & 6 & 3 & 3 & 2 & 4 & 1\\
 %& and Indian Bowl & & &  &  &  &  &  & \\
%August Macke & Big Zoo, Triptych & 1 & 10 & 0 & 0 & 1 & 3 & 1 & 0\\
%Gustav Klimt & Ode To Klimt & 12 & 4 & 2 & 3 & 2 & 1 & 3 & 0\\
%Heinrich Campendonk & Cyclist E. and little yellow cow & 2 & 4 & 0 & 0 & 1 & 1 & 0 & 1\\
%Jean Metzinger & Corkscrew glass and matches  & 5 & 1 & 0 & 1 & 4 & 0 & 1 & 0\\
%Max Ernst & The Antipope & 0 & 0 & 0 & 0 & 2 & 0 & 1 & 0\\
%Max Pechstein & Sunset  & 0 & 0 & 0 & 2 & 0 & 0 & 0 & 0\\
%    \bottomrule
 % \end{tabular}
%\end{table*}

In the following, we investigate a selection of these 21 paintings. The paintings \textit{Banks of Seine}\footnote{\url{https://www.wikiart.org/en/andre-derain/banks-of-seine} (\lastvisiteddatum).} 
or \textit{London Bridge}\footnote{\url{https://www.wikiart.org/en/andre-derain/london-bridge-1906} (\lastvisiteddatum).} and \textit{Still Life with Pears and Indian Bowl} 
% \footnote{https://www.wikiart.org/en/andre-derain/still-life-with-pears-and-indian-bowl (\lastvisiteddatum)).}
\footnote{\url{https://www.sothebys.com/en/buy/auction/2021/collection-de-monsieur-et-madame-robert-schmit-oeuvres-choisies-session-i/nature-morte-aux-poires} (\lastvisiteddatum).} are currently all attributed to André Derain and were all detected by our 7 Efficientnet models and KAN as having multiple disputable patches. The same accounts for the painting \textit{Ode To Klimt} attributed to Gustav Klimt which also had at least one highly disputable patch detected by any of our 7 complexity models. In the case of \textit{Big Zoo, Triptych}\footnote{\url{https://www.wikiart.org/en/august-macke/big-zoo-triptych} (\lastvisiteddatum).} attributed to August Macke, \textit{Cyclist E. and little yellow cow}\footnote{\url{https://www.kunsthaus.nrw/sammlungen/heinrich-campendonk-2} (\lastvisiteddatum).} attributed to Heinrich Campendonk or \textit{Corkscrew glass and matches}\footnote{\url{https://www.christies.com/en/lot/lot-5938586} (\lastvisiteddatum).} attributed to Jean Metzinger, at least one highly disputable patch was detected by 4 of our 7 complexity models.

\begin{figure}[tb]
  \centering
    \begin{subfigure}{0.49\linewidth}
  \includegraphics[width=6cm]{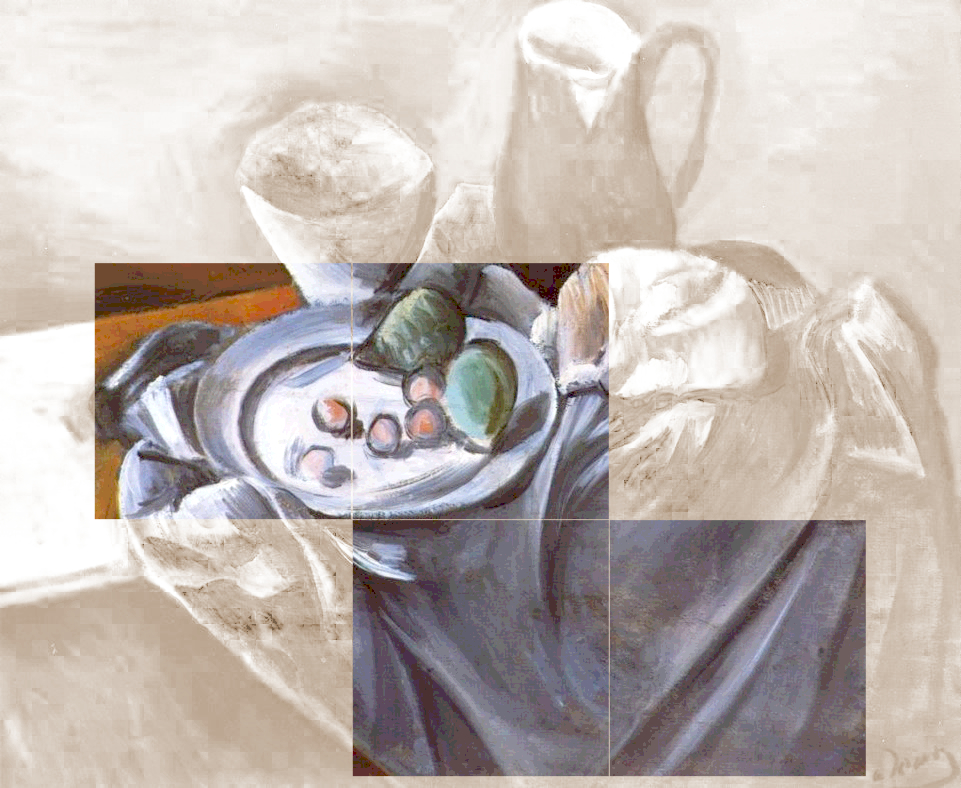}
%    \fbox{\rule{0pt}{0.5in}    \rule{.9\linewidth}{0pt}}
    \caption{Patch in the low-resolution}
    \label{figure:Andre-Derain_47231_Patches-a}
  \end{subfigure}
  \hfill
  \begin{subfigure}{0.49\linewidth}
  \includegraphics[width=6.1cm]{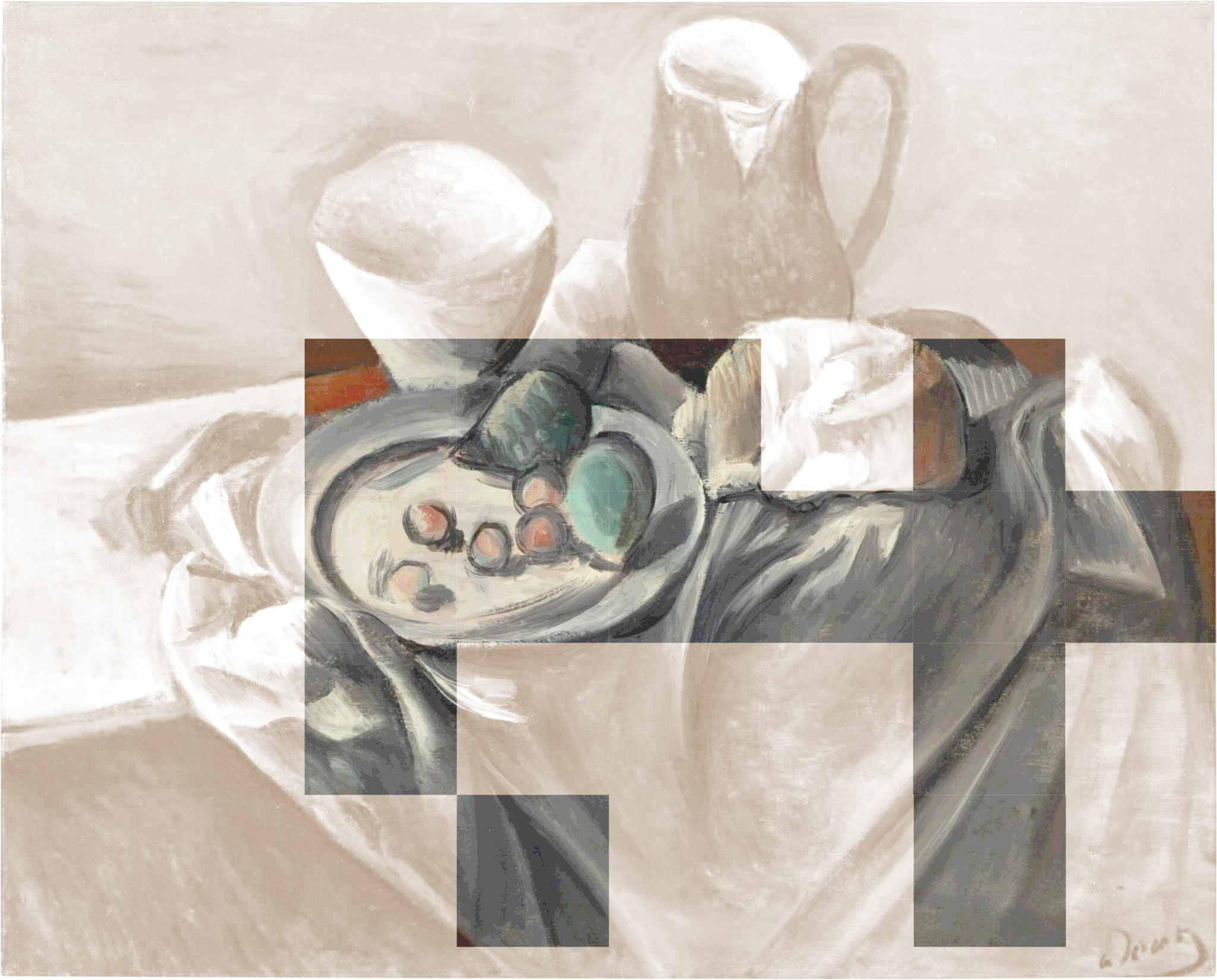}
    %\fbox{\rule{0pt}{0.5in}  \rule{.9\linewidth}{0pt}}
    \caption{Patch in the high-resolution}
    \label{figure:Andre-Derain_47231_Patches-b}
  \end{subfigure}
  \caption{Image of the painting "Still Life with Pears and Indian Bowl" attributed to André Derain that where attributed to Wolfgang Beltracchi by 4 of our 7 Models trained with $T38$. Image in the public domain.}
  \label{figure:Andre-Derain_47231_Patches}
\end{figure}

\begin{figure}[tb]
  \centering
    \begin{subfigure}{0.49\linewidth}
  \includegraphics[width=6cm]{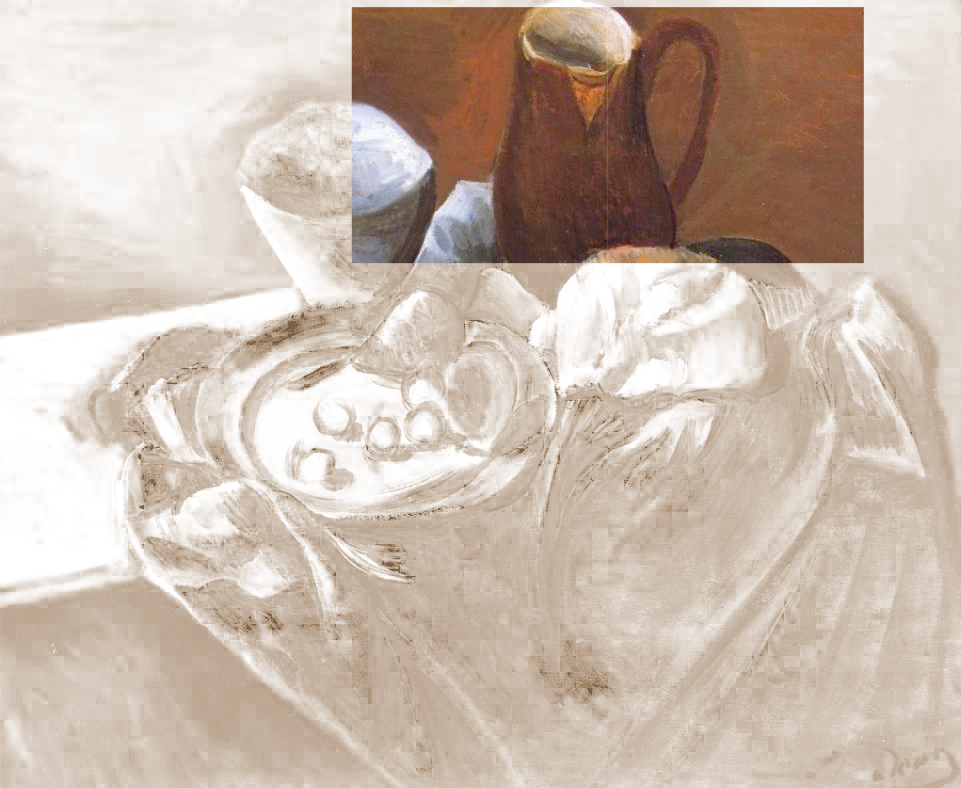}
      \caption{Patch in the low-resolution}
    \label{figure:Andre-Derain_47231_Patches_Kan-a}
  \end{subfigure}
  \hfill
  \begin{subfigure}{0.49\linewidth}
  \includegraphics[width=6.1cm]{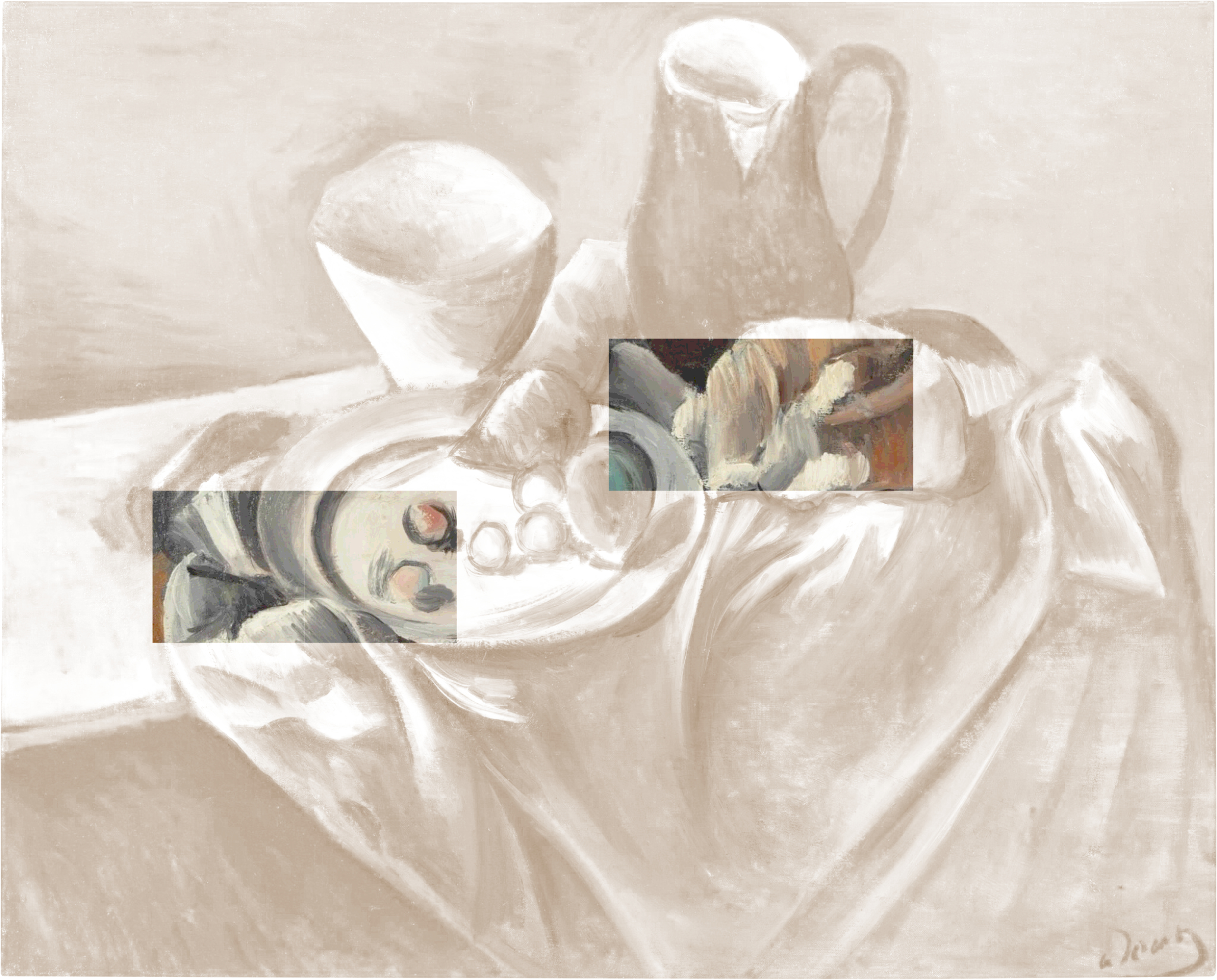}
    %\fbox{\rule{0pt}{0.5in}  \rule{.9\linewidth}{0pt}}
    \caption{Patch in the high-resolution}
    \label{figure:Andre-Derain_47231_Patches_Kan-b}
  \end{subfigure}
  \caption{Image of the painting "Still Life with Pears and Indian Bowl" attributed to André Derain that where attributed to Wolfgang Beltracchi by KAN Model (Width 120/84/12, Grid = 5, k = 3) trained with $T38$. Image in the public domain.}
  \label{figure:Andre-Derain_47231_Patches}
\end{figure}

To further validate our findings, we test which patches are predicted as disputed on images with different resolutions. We show the results for Andrè Derrain's \textit{Still Life with Pears and Indian Bowl} in \Cref{figure:Andre-Derain_47231_Patches-a}. We see that even in the higher-resolution image our models detected similar areas in the painting as Beltracchi. Interestingly, none of the patches in the painting was attributed to André Derain. %\Cref{figure:DerainStillLife} shows the comparison of the patch attribution towards André Derain or Wolfgang Beltracchi only.

%We now only considered patches where the model predicted a negative value to the official known attribution. Meaning that the model did rather not attributed the particular patch to the artist.
% And we further 

%\begin{figure*}
 % \centering
  %  \includegraphics[width=10cm]{images/Andre-Derain_image_to_test_entropy_2_5_seed_38_b5_calculate_predictions}
  %\caption{Distribution of patches from the painting "Still Life with Pears and Indian Bowl"  attributed to André Derain evaluated by model B5.}
   %\label{figure:DerainStillLife}
%\end{figure*}

\subsection{Art Historical Interpretation}
\label{sec:artHistoricalInterpretation}

To understand the process of forgery detection in our algorithm, we contextualize its results from an art historical perspective. This is done by conducting a visual analysis to identify the stylistic features primarily detected by the algorithm in its attribution of works to Wolfgang Beltracchi. Three paintings function as the focus of this analysis, including: \textit{Banks of Seine}, \textit{London Bridge}, and \textit{Still Life with Pears and Indian Bowl}. While bearing official authentication as works by André Derain, multiple patches of these works have been detected as Beltracchi by each model of our algorithm. As such, these works depict the elements most confidently linked to Beltracchi, so much as to arouse suspicion of genuine works. This allows the visual elements most likely detected as Beltracchi to be singled out. Ultimately, these inform our understanding of the dataset and its biases, driving further development of the algorithm for application to the art market.

%In the following, we refer to observations from the dataset $T38$ trained on EfficientNet B0 with an Entropy > 2.5.

\subsubsection{Visual Analysis}
The stylistic features of André Derain's \textit{Banks of Seine}, \textit{London Bridge}, and \textit{Still Life with Pears and Indian Bowl} are analyzed according to their available metadata of style and genre.

André Derain is known for working in the Fauvist style. As such, these works show a particularly high degree of painterly qualities. This is visible in their distinguishable and spontaneous use of brushstrokes and light. In \textit{Still Life with Pears and Indian Bowl}, the patches attributed to Beltracchi center around the section of the tablecloth lying on top of the table, including the plate of fruits (Figure \ref{figure:Andre-Derain_47231_Patches}). The spontaneous application of brushstrokes in these patches confers a sense of motion. Those outlining the fruits, for instance, seemingly convey a kinetic quality to the still objects. Notably, the fruits represent the highest attribution to Beltracchi and lowest to Derain. From a source off on the left side of the painting, light is directed onto the same section, reflected in the white highlights on the tablecloth and shadows behind the fruits. This introduction of light directs the viewer's vision along the highly Beltracchi-attributed patches.
    
In \textit{London Bridge}, this emphasis on motion and light is particularly visible on the water. The arrangement of short but wide brushstrokes conveys a sense of spontaneity that captures our line of vision, primarily in the yellow brushstrokes aligned vertically along the left third of the picture plane. The reflection of sunlight scattering off of the water conveys a sense of motion.
    
In the \textit{Banks of Seine}, directional light is not clearly implemented, yet the elongated and curved brushstrokes of the pavement convey a sense of speed that warps the viewer's vision, as if moving along the river. Stylistically, the attribution of Beltracchi to these works by Derain may thus be captured in the strong emphasis on motion and light, implemented by the Fauvist application of visible brushstrokes. These findings are supported by Beltracchi's works in our dataset, being well-representative of kinetic elements, regardless of style. A Fauvist emphasis on color is not specifically emphasized in these works.

Two genres are represented among the relevant works, including two landscapes (\textit{Banks of Seine} and \textit{London Bridge}) and a still life painting (\textit{Still Life with Pears and Indian Bowl}). In their representation of scenes, as opposed to more abstract or figurative motifs, these works further effectuate a clear line of vision. In both landscapes, the viewer’s eyes are directed along the top left and bottom right corners of the paintings, along the bridge in \textit{London Bridge} and the road in \textit{Banks of Seine}. The same effect is achieved in the \textit{Still Life with Pears and Indian Bowl}, yet in the opposite direction, following the tabletop along the center of the composition. While uncharacteristic of Fauvist painting, such movement of vision to opposite ends of the picture planes creates a clear fore, middle, and background. Respectively, these constitute the large boat, bridge, and city in \textit{London Bridge}, the road, nearest figure, and distant city in (\textit{Banks of Seine}, and finally the section under the diagonal edge of the tabletop, the tabletop with foods and bowls, and the back wall, in \textit{Still Life with Pears and Indian Bowl}. As such, the detection of Beltracchi in these works may relate to their implementation of a clear line of vision, resulting in compositional depth. While less apparent among other works with Beltracchi-attributed patches, a clear dimensional division remains visible among them. The Beltracchi works in our dataset are well-representative of landscape painting, yet less of still lives, in favor of kinetic elements.

The noteworthy prevalence of patches attributed to Wolfgang Beltracchi in works by André Derain implies that Fauvist landscapes and still lifes exemplify features linked to Beltracchi. This occurs primarily according to their spontaneous use of brushstrokes. While characteristic of Fauvist painting, this style is not well-represented across our dataset (Figure \ref{table:artistInDataset}). Other Fauvist artists, such as Jean Metzinger and Maurice de Vlaminck find even less representation than Derain. The spontaneity of the brushstrokes in these works, however, stylistically translates across modern art movements. The use of short brushstrokes, as emphasized in \textit{London Bridge}, is particularly characteristic of the Impressionist work of Vincent van Gogh, best represented in our dataset. While the genres of landscape and still life painting are represented in the work of van Gogh, many works by Beltracchi in our dataset are figurative works in cubist style. As Beltracchi is known for forging various artistic styles, his forgeries thus ought to be evaluated according to his exact handling of the paint. As can be deduced from our algorithm, Beltracchi tends to forge works with a clear implementation of light, motion, and line of vision. By implementing a closer analysis of Beltracchi's brushwork, precise details in his handling of these features may become clear.

\section{Critical Discussion}
\label{sec:criticalDiscussion}

 The application of Artificial Intelligence to the task of artistic forgery detection shows high accuracy. It is important to note that the accuracy of the models is strictly dependent upon the complexity of the dataset.  On the basis of our dataset, the algorithm has learned to attribute artists to specific works, including those created by the forger Wolfgang Beltracchi. The strengths of this approach are in providing an efficient and objective method for forgery detection. Combining this method with art historical knowledge, however, further contextualizes the predictions and enables their understanding based on specific stylistic features. As such, the combined approach to AI and art history creates mutual support. Critical analysis of algorithmic predictions and the continual improvement and extension of the algorithm should be included to strengthen the theoretical foundation.

\section{Conclusion \& Future Work}
\label{sec:conclusions}
%In this paper, we contribute toward art authentication research in showing that challenging paintings towards famous forgers can help detect inconsistencies faster with the help of artificial intelligence techniques. 
%Such discrepancies enable art experts to concentrate on certain parts of a painting where the attribution to an artist is rather questionable or where a forger even reveals a mistake that has been made.
%And if needed other potentially more costly authentication techniques can thus be quickly applied directly in appropriate places.
%We showed that our models trained on top of EfficientNet\cite{tan2019efficientnet} can attribute paintings done by a forger to a forger even if done in the style of a famous artist with a reasonable degree of accuracy. Lastly, we compared different EfficientNet models towards their performance in the image classification and attribution domain to get further insides to how deep networks should be to perform best in such tasks.

In this paper, we contribute toward art authentication research by showing the potential of CNN and KAN models for challenging paintings of famous forgers and detecting inconsistencies. 
Such discrepancies enable art experts to concentrate on certain parts of a painting where the attribution to an artist is questionable.

We showed that our models trained on top of EfficientNet\cite{tan2019efficientnet} can detect paintings in the style of a famous artist with a reasonable degree of accuracy. Second, we compared different EfficientNet models to determine which model size is most effective and compared the results to the newly introduced KAN models. Lastly, contextualization of the results from an art historical perspective has been shown according to style and subject matter to differentiate the visual analysis. In determining a painting as forged by Beltracchi, a comparison of patches in works of the same style reveals that artworks may be more readily attributed to Beltracchi according to more precise features of light, motion, and composition. When considering the subject matter, Beltracchi’s work tends to be linked to classical motifs of landscape and still life painting. In addition to further stylistic elements, forgery is determined with greater certainty.

Further work should be dedicated to finding the optimal KAN architecture and training procedure to more fairly compare the results to traditional convolutional neural networks. Moreover, in future work, we wish to test additional methods to mitigate training set mislabeling, including using an additional model to first predict possibly mislabeled data and train excluding the flagged data.

% \noindent 

The next steps include extending the datasets with more high-resolution pictures and cross-evaluating the results with art experts.
Further, we wish to train models on other famous forgers, apply the techniques to a larger number of artworks, and eventually work side by side with a museum curator or an auction house to gather more specific insights.

% \begin{figure}[tb]
%   \centering
%   \begin{subfigure}{0.68\linewidth}
%    \fbox{\rule{0pt}{0.5in} \rule{.9\linewidth}{0pt}}
%     \caption{An example of a subfigure}
%     \label{fig:short-a}
%   \end{subfigure}
 %  \hfill
%   \begin{subfigure}{0.28\linewidth}
%     \fbox{\rule{0pt}{0.5in} \rule{.9\linewidth}{0pt}}
%     \caption{Another example of a subfigure}
%     \label{fig:short-b}
%   \end{subfigure}
%   \caption{Centered, short example caption}
%   \label{fig:short}
% \end{figure}

\section*{Acknowledgements}
Special thanks go to Pepe Ballesteros Zapata and Darío Negueruela del Castillo from the Digital Visual Studies Group at the University of Zurich (\url{https://dvstudies.net}) for their support during the seminar that led to this work.

\newpage
\bibliographystyle{eccv2024style/splncs04}
\bibliography{main}

\end{document}